\documentclass{article}
\usepackage{spconf,amsmath,graphicx}
\usepackage{amsmath}
\usepackage{amsfonts,amssymb} 
\usepackage{float}
\usepackage{color}
\usepackage{multirow}
\usepackage{booktabs}

\title{Recognition-Guided Diffusion Model for Scene Text Image Super-Resolution}
\name{Yuxuan Zhou, Liangcai Gao, Zhi Tang, Baole Wei}
\address{Wangxuan Institute of Computer Technology, Peking University, Beijing, China}
\begin{document}
%
\maketitle
\begin{abstract}
Scene Text Image Super-Resolution (STISR) aims to enhance the resolution and legibility of text within low-resolution (LR) images, consequently elevating recognition accuracy in Scene Text Recognition (STR). Previous methods predominantly employ discriminative Convolutional Neural Networks (CNNs) augmented with diverse forms of text guidance to address this issue. Nevertheless, they remain deficient when confronted with severely blurred images, due to their insufficient generation capability when little structural or semantic information can be extracted from original images. Therefore, we introduce RGDiffSR, a Recognition-Guided Diffusion model for scene text image Super-Resolution, which exhibits great generative diversity and fidelity even in challenging scenarios. Moreover, we propose a Recognition-Guided Denoising Network, to guide the diffusion model generating LR-consistent results through succinct semantic guidance. Experiments on the TextZoom dataset demonstrate the superiority of RGDiffSR over prior state-of-the-art methods in both text recognition accuracy and image fidelity.

\end{abstract}
\begin{keywords}
scene text image super-resolution, diffusion model,
attention mechanism, scene text recognition
\end{keywords}

\section{Introduction}
Scene Text Recognition (STR) adopts computer vision and machine learning methods\cite{ABINet,PARSeq} to recognize individual characters or words in a range of diverse and intricate scenarios, and has found utility across various practical applications. However, this technique remains susceptible to the limitations posed by low-resolution (LR) images, which may engender erroneous recognition results due to image blurring or spatial distortion.

\begin{figure}[htbp]
\centering
\includegraphics[width=0.9\linewidth]{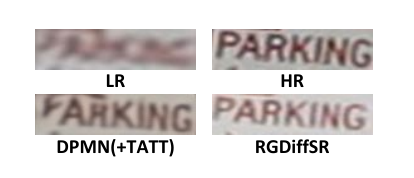}
\caption{SR images generated by DPMN+TATT (bottom-left) and RGDiffSR (bottom-right) from heavily blurred LR image.}
\label{intro}
\end{figure}

In recent years, Scene Text Image Super-Resolution (STISR) has emerged as a pivotal area of research, aiming to enhance the quality and clarity of scene text images. In the early stage, researchers employed conventional Single Image Super-Resolution (SISR) methods \cite{SRCNN} on manually downsampled datasets of scene text images. To further enhance the quality of generated text region, the Text Super-Resolution Network (TSRN)\cite{TSRN} is introduced alongside a specialized STISR dataset named TextZoom. 
Recent state-of-the-art approaches \cite{gestalt,TATT,C3,DPMN} mostly use diverse forms of text-based guidance such as text mask and recognition results to furnish semantic-level information to TSRN backbone. Yet, despite the growing complexity of their guidance mechanisms, these approaches still fall short in generating satisfactory results when confronted with extremely challenging scenarios such as heavily blurred images, as depicted in Figure \ref{intro}.

Acknowledging the inherent limitations posed by the generation capability of TSRN backbones in the aforementioned methods, we propose RGDiffSR, based on generative diffusion model.
\begin{figure*}[htbp]
\centering
\includegraphics[width=\linewidth]{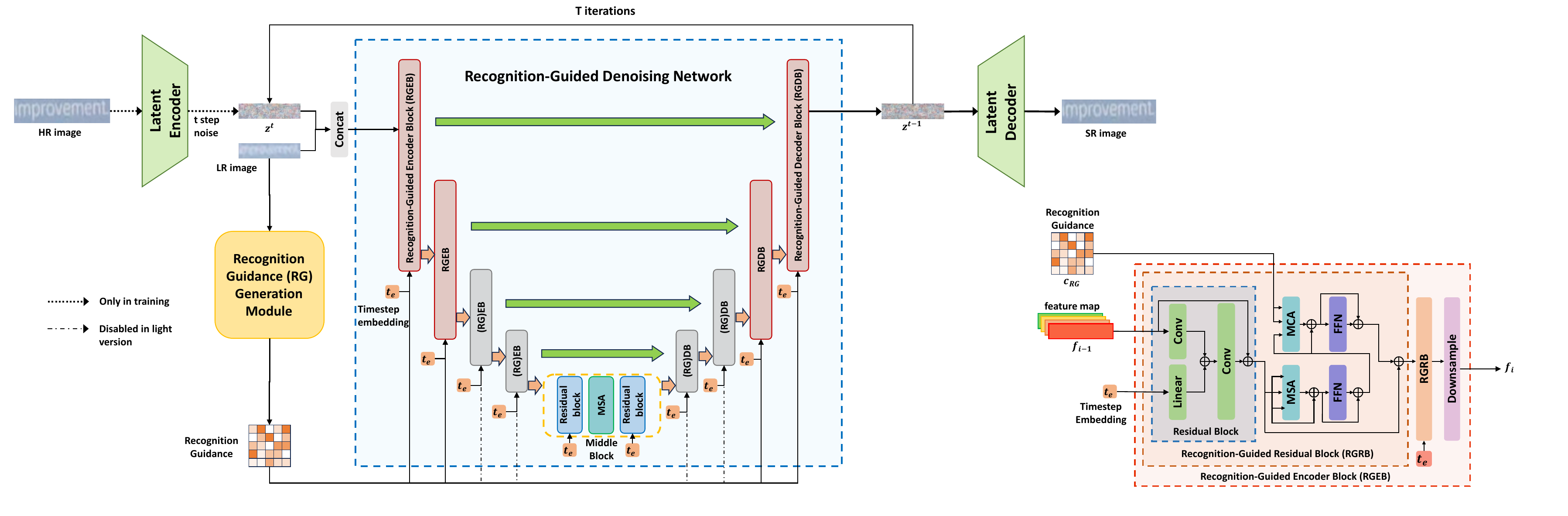}
\caption{The overall architecture of RGDiffSR and the structure of Recognition-Guided Encoder Block (bottom-right).}
\label{all}
\end{figure*}
Diffusion models have emerged as the forefront of generative models recently, not only achieving SOTA performance in general SISR tasks\cite{SR3,SRDiff}, but also exhibiting notable power in integrating cross-modal information in large text-to-image diffusion models such as GLIDE \cite{GLIDE} and Imagen \cite{Imagen}. 
By employing the diffusion sampling process, RGDiffSR generates text images exhibiting heightened fidelity and diversity, even in challenging scenarios. Furthermore, to guide the model recovering text regions with more distinctive character shapes, we introduce a Recognition-Guided Denoising Network. This novel component leverages both the low-resolution image and recognition results from a STR recognizer as conditions within the diffusion process, merging semantic guidance with pixel-level information through an attention mechanism.


Our contributions can be summarized as follows: (1) We
propose RGDiffSR, which is the first work applying diffusion-based generative network to STISR task, to the best of our knowledge. (2) The proposed Recognition-Guided Denoising Network provides noise prediction generated with semantic guidance in diffusion process, resulting in a substantial improvement in the legibility of text regions. (3) Experiment results show that RGDiffSR surpasses existing STISR methods largely in terms of recognition accuracy without compromising fidelity.

\section{Methodology}
\subsection{Model Overview}

Given low-resolution image $x_{LR}\in \mathbb{R}^{H\times W\times C}$, where $H,W,C$ represents height, width and channels respectively, the goal of STISR is to generate high-resolution images $x_{SR}\in \mathbb{R}^{fH\times fW\times C}$ that enhance legibility for both human and artificial recognizers. Here $f$ signifies the scale factor. The overall architecture of RGDiffSR is shown in fig \ref{all}.
We adopt the Latent Diffusion Model as the basic super-resolution pipeline to reduce the computational complexity while retaining the generative capability of diffusion models. During training stage, the model is given LR-HR image pair $(x_{LR},x_{HR}^k\in \mathbb{R}^{fH\times fW\times C})$. The HR image $x_{HR}$ is first compressed by the latent encoder $\mathcal{E}$ into latent representation $z\in \mathbb{R}^{H\times W\times C}$, which share the same dimensions with LR image, then be added with a series of Gaussian noises, eventually turning into $z_T$ in Gaussian distribution. Meanwhile, the LR image is fed into the recognition guidance generation module to extract the recognition guidance $c_{RG}$, which will be sent into the diffusion model along with the LR images as condition, to guide the denoising network recover the original latent $z_0$ from Gaussian vector $z_{T}$. Finally, the SR image $x_{SR}$ is reconstructed through latent decoder $\mathcal{D}$ from $z_0.$

\subsection{Diffusion models}
The latent encoder/decoder is a VQ-regularized\cite{VQGAN} 2x autoencoder trained separately following \cite{LatentDiffusion}, compressing the HR image to a perceptually equivalent latent space to lower the computational demands.
Then the diffusion model can be decomposed into 2 processes: the forward diffusion process and the reverse process. In forward process, a sequence of small Gaussian noise is gradually added to the latent vector $z$ for $T$ steps. The sizes of noise are controlled by a variance schedule $\{\beta_t\}_{t=1}^T$. Suppose that the latent data distribution is $q(z)$, this forward process can be formulated as:
\begin{align}
q(z_t|z_{t-1})&=\mathcal{N}(z_t;\sqrt{1-\beta_t}z_{t-1},\beta_t\bf{I})
\end{align}
By using some reparameterization tricks, $z_t$ at any timestep $t$ can be sampled through:
\begin{align}
q(z_t|z_0)=\mathcal{N}(z_t;\sqrt{\bar{\alpha_t}}z_0,(1-\bar{\alpha_t})\bf{I})
\end{align}
where $\alpha_t=1-\beta_t,\bar{\alpha}_t=\prod_{k=1}^t\alpha_k.$ Eventually, the latent vector would resemble the standard Gaussian:
\begin{align}
q(z_T|z_0)=\mathcal{N}(z_T;\bf{0},\bf{I})    
\end{align}

Since the $\beta_t$ is small enough, the distribution of reverse step $q(z_{t-1}|z_t)$ will also be Gaussian. Therefore the reverse step $p_{\theta}(z_{t-1}|z_t)$ can be modeled through a neural network $\theta$, which corresponds to the denoising network. In super-resolution tasks, diffusion models generate corresponding HR images by taking the LR images as condition, and the conditional form of reverse process can be written as:
\begin{align}
p_{\theta}(z_{t-1}|z_t,x_{LR})=\mathcal{N}(z_{t-1};\mu_{\theta}(z_t,x_{LR},t),\Tilde{\beta_t}\bf{I}),    \\
\mu_{\theta}(z_t,x_{LR},t)=\dfrac{1}{\sqrt{\alpha_t}}(z_t-\dfrac{1-\alpha_t}{\sqrt{1-\bar{\alpha_t}}}\epsilon_{\theta}(z_t,x_{LR},t)),    
\end{align}

where $\Tilde{\beta}_t=\frac{1-\bar{\alpha}_{t-1}}{1-\bar{\alpha}_t}\beta_t$, $\epsilon_{\theta}$ is the noise predicted by neural network, and the mean $\mu_{\theta}$ is interpreted through Bayes rules.

In training phase, we optimize the variational lower bound of negative log-likelihood $\mathbb{E}[-\log p_\theta(z_0)]$, getting the L2-form loss function:
\begin{align}
L_{DM}=\mathbb{E}_{\mathcal{E}(x_{HR}),\epsilon\sim \mathcal{N}(0,1),t}[\|\epsilon-\epsilon_\theta(z_t,x_{LR},t)^2\|]    
\end{align}

In inference phase, we first sample the latent vector from a random Gaussian noise through the reverse process:
\begin{align}
z_{t-1}=\dfrac{1}{\sqrt{\alpha_t}}(z_t-\dfrac{1-\alpha_t}{\sqrt{1-\bar{\alpha_t}}}\epsilon_{\theta}(z_t,x_{LR},t))+\sqrt{\beta_t}\epsilon_t    ,
\end{align}

where $\epsilon_t\sim \mathcal{N}(\bf{0},\bf{I}).$ Then the recovered latent vector $z_0$ is decoded to original pixel space by the latent decoder, generating the SR image ${x}_{SR}=\mathcal{D}(z_0).$


\subsection{Recognition Guidance Generation Module}
The Recognition Guidance Generation Module (RGGM) is a STR recognizer, which takes the LR image as input and predicts the probability distribution of which class each character belongs to. Thus the recognition guidance can be denoted as: $c_{RG}=\mathcal{R}(x_{LR})\in \mathbb{R}^{L\times |\mathcal{A}|}$, where $L$ denotes the max predict length, and $|\mathcal{A}|$ is the size of the character set.
Considering the significance of recognition guidance in enhancing the quality of text regions within the SR image, elevating the recognition accuracy of the recognizer is noticeably beneficial. Consequently, during the training phase, the recognizer is fine-tuned by a text recognition loss:
\begin{align}
L_{recog}=\|\mathcal{R}(x_{LR})-\mathcal{R}(x_{HR})\|_1    
\end{align}
\subsection{Recognition-Guided Denoising Network}

The recognition-guided denoising network is based on U-Net with the attention mechanism, as depicted in Fig \ref{all}.
Following the typical U-Net architecture, the denoising network consists of 4 encoder blocks, 1 middle block, and 4 decoder blocks. Each block $b_i$ takes the feature $f_{i-1}$ from the former block and the timestep embedding $t_e$ as input, where $t_e$ is embedded through a sinusoidal positional encoding, and $f_0$ is the concatenation of $z_T$ and $x_{LR}$.


Specifically, in each Recognition-Guided Encoder/Decoder Block (RGEB/RGDB) $B_i$, the image feature $f_{i-1}$ will go through 2 recognition-guided residual block (RGRB) and a downsample/upsample layer. In RGRB, $f_{i-1}$ is first fused with the timestep embedding $t_e$ through an element-wise sum in the common residual block. Subsequently, the output $h^{res}_{i,0}$ is fed into a Multihead Self Attention (MSA) layer that employs dot-product attention to capture global correlations between pixels.
After passing layer-norm layers(omitted in Fig \ref{all}) and a Feed Forward Network(FFN), the feature $h^{MSA}_{i,0}$ is sent to Muitihead Cross Attention(MCA) layer to absorb the semantic  from recognition guidance $c_{RG}.$ The whole process in RGEB can be written as:
\begin{align}
h^{res}_{i,0}&=Res(f_{i-1},t_e) \label{r1} \\
h^{MSA}_{i,0}&=FFN(LN(MSA(h^{res}_{i,0}))) \label{r2} \\
h^{MCA}_{i,0}&=FFN(LN(MCA(h^{MSA}_{i,0},c_{RG},c_{RG}))) \label{r3} \\
h^{MCA}_{i,1}&=RGRB_1(h^{MCA}_{i,0},t_e,c_{RG})  \\
f_i&=Downsample(h^{MCA}_{i,1}) 
\end{align},
where Eq. \ref{r1},\ref{r2},\ref{r3} compose the first residual block $RGRB_0$.

To further extract the global relation, an MSA layer is also applied between 2 residual blocks in the middle block. The configurations in decoder blocks mirror those in encoder blocks, with the only alteration of replacing Downsample layers with upsample layers.

The attention design empowers the denoising network to effectively leverage semantic information contained in text regions, consequently generating images that demonstrate more distinct character shapes.
\begin{table*}[htbp]
  \centering
  \resizebox{\linewidth}{!}{
    \begin{tabular}{l|c c c c|c c c c|c c c c}
      \hline
      & \multicolumn{4}{c}{ASTER\cite{ASTER}}& \multicolumn{4}{c}{MORAN\cite{MORAN}}& \multicolumn{4}{c}{CRNN\cite{CRNN}}\\
      \hline
      Method & easy & medium & hard & average & easy & medium & hard & average & easy & medium & hard & average\\
      \hline
      bicubic & 64.7\% & 42.4\% & 31.2\% & 47.2\% & 60.6\% & 37.9\% & 30.8\% & 44.1\% & 36.4\% & 21.1\% & 21.1\% & 26.8\%\\
      \hline
      $\text{TG}_\text{AAAI2022}$\cite{gestalt} & 77.9\% & 60.2\% & 42.4\% & 61.3\% & 75.8\% & 57.8\% & 41.4\% & 59.4\% & 61.2\% & 47.6\% & 35.5\% & 48.9\%\\
      $\text{TATT}_\text{CVPR2022}$\cite{TATT} & 78.9\% & 63.4\% & 45.4\% & 63.6\% & 72.5\% & 60.2\% & 43.1\% & 59.5\% & 62.6\% & 53.4\% & 39.8\% & 52.6\%\\
      $\text{C3-STISR}_\text{IJCAI2022}$\cite{C3} & 79.1\% & 63.3\% & 46.8\% & 64.1\% & 74.2\% & 61.0\% & 43.2\% & 60.5\% & 65.2\% & 53.6\% & 39.8\% & 53.7\%\\      
      $\text{DPMN(+TATT)}_\text{AAAI2023}$\cite{DPMN} & 79.3\% & 64.1\% & 45.2\% & 63.9\% & 73.3\% & 61.5\% & 43.9\% & 60.4\% & 64.4\% & 54.2\% & 39.2\% & 53.4\%\\        $\text{TSAN}_\text{AAAI2023}\cite{TSAN}$ & 79.6\% & 64.1\% & 45.3\% & 64.1\% & 78.4\% & 61.3\% & 45.1\% & 62.7\% & 64.6\% & 53.3\% & 38.8\% & 53.0\%\\
      \hline
      RGDiffSR & \textbf{81.1}\% & \textbf{65.4}\% & \textbf{49.1}\% & \textbf{66.2}\% & \textbf{78.6}\% & \textbf{62.1}\% & \bf{45.4}\% & \textbf{63.1}\% & \textbf{67.6}\% & \textbf{56.5}\% & \textbf{42.7}\% & \textbf{56.4}\%\\
      \hline
      HR & 94.2\% & 87.7\% & 76.2\% & 86.6\% & 91.2\% & 85.3\% & 74.2\% & 84.1\% & 76.4\% & 75.1\% & 64.6\% & 72.4\%\\
      \hline
    \end{tabular}
  }
  \caption{Comparison of downstream text recognition accuracy with other SOTA methods on TextZoom dataset. Bolded numbers denote the best results.}
  \label{accuracy}
\end{table*}
\section{Experiments}

\subsection{Experimental Setup}
RGDiffSR is implemented in pytorch 2.0.1, trained with an AdamW optimizer for 250 epochs. The batch size is 64, and the learning rate is set to 1e-6. The channel size of the feature in encoder block is multiplied by 1,2,2,4 in order, with 160 as the initial size. The number of heads in each MSA and MCA layer is 8. In training, the total timesteps of diffusion process is set to 1000, and the noise weight $\{\beta_t\}_{t=1}^T$ is scheduled following \cite{noise}. In sampling, a DDIM\cite{DDIM} sampler is used to accelerate the reverse process, with 200 DDIM steps. 
CRNN\cite{CRNN} is used as the STR recognizer in recognition guidance generation module.

\subsection{Dataset}
All experiments are conducted on the TextZoom\cite{TSRN} dataset, which contains 21,740 LR-HR text image pairs, captured in diverse real-world scenarios using cameras with varying focal lengths. 4,373 pairs of them are divided into three distinct test subsets based on their recognition difficulty, namely easy (1,619 pairs), medium (1,411 pairs), and hard (1,343 pairs). The size of LR and HR images are standardized to $16\times64$ and $32\times 128$ respectively.


\subsection{Comparison to State-of-the-Arts}

\begin{figure}[ht]
\centering
\includegraphics[width=\linewidth]{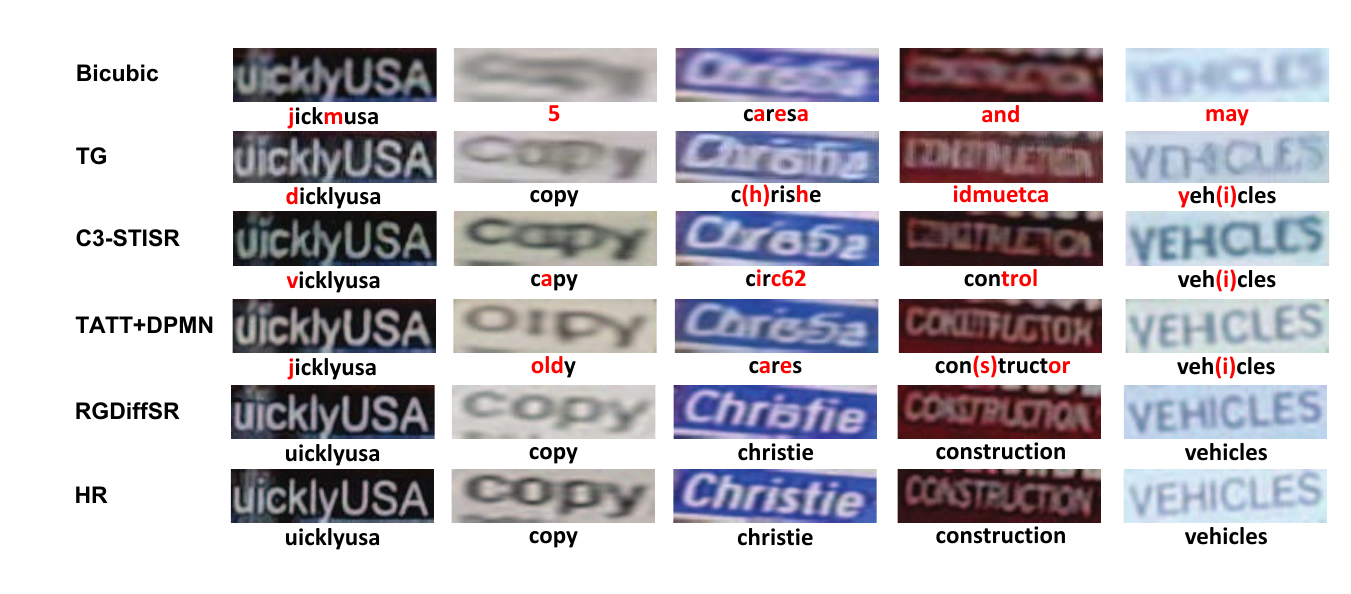}
\caption{Qualitative comparison with other SOTA methods.}
\label{compare}
\end{figure}

The recognition accuracy presented in Table \ref{accuracy} emphasizes the superior capabilities of RGDiffSR across various situations compared to previous SOTA method. Notably, our approach performs even better in hard cases, displaying a substantial accuracy improvement of 3.8\% and 3.9\% respectively when employing ASTER and CRNN as recognizers. 
Additionally, the visual representation of the generated SR images, as depicted in Figure \ref{compare}, demonstrates that RGDiffSR also achieves impressive fidelity and legibility in generated images.
\subsection{Effectiveness of Recognition-Guided Denoising Network}\label{eff}
\begin{table}[htbp]
  \centering
  \resizebox{\linewidth}{!}{
      \begin{tabular}{l|c c c c|c}
        \hline
         \multirow{2}{*}{RG block ids} &\multicolumn{4}{c|}{recognition accuracy} &\multirow{2}{*}{number of parameters}\\      
        \cline{2-5}
        & easy & medium & hard & average \\
        \hline
        $\emptyset$ & 74.6\% & 58.6\% & 41.2\% & 59.2\% & 135.3M\\ 
        \{1,2\} (light ver.) & 81.6\% & 64.9\% & 48.9\% & 66.1\% & 148.3M\\  
        \{3,4\} & 77.8\% & 60.4\% & 44.3\% & 61.9\% & 186.9M\\ 
        \{1,2,3,4\} & 81.1\% & 65.4\% & 49.1\% & 66.2\% & 200.0M\\ 
        \hline
      \end{tabular}
  }
  \caption{Average recognition accuracy on ASTER and parameter amount when different Encoder/Decoder Blocks are recognition-guided.}
  \label{ablation}
\end{table}


To verify the effectiveness of Recognition-Guided Denoising Network, we perform experiments on U-net with different Encoder/Decoder Blocks being guided with recognition results. 
As shown in Table \ref{ablation}, when all Encoder/Decoder blocks are recognition-guided, the average accuracy is largely boosted by 7\%.
Intriguingly, when only the first 2 Encoder Blocks and the last 2 Decoder Blocks are recognition-guided, the result only exhibits a marginal drop of 0.1\% in comparison to the fully guided version. 
This observation strongly implies that recognition guidance more effectively interacts with shallow feature maps, introducing the light version of RGDiffSR, which not only preserves accuracy but also reduces parameter amount by 25\%.

\subsection{Trade-off between Fidelity and Accuracy}

\begin{table}[htbp]
  \centering
  \resizebox{\linewidth}{!}{
      \begin{tabular}{l|c c c c|c c c}
        \hline
        method & & PSNR & SSIM & & &accuracy &\\
        \hline
        bicubic & &  20.35 & 0.6961 & & &47.2\% &\\        
        \hline
        TATT & & 21.52 & 0.7930 & & &63.6\%  &\\  
        C3-STISR & & 21.60 & \color{blue}{0.7931} & & &64.1\%  &\\  
        DPMN(+TATT)& & 21.49 & 0.7925& & &63.9\% & \\ 
        TSAN& & \color{red}{22.10} & 0.7835 & & &64.1\%  &\\  
        \hline
        RGDiffSR-250& & 21.31 & 0.7582 & & &\color{red}{66.2}\% & \\        
        RGDiffSR-1000& & \textcolor{blue}{21.88} & \color{red}{0.7962}& & &\color{blue}{65.9}\% &\\        
        \hline
      \end{tabular}
  }
  \caption{PSNR/SSIM and average recognition accuracy on ASTER of different STISR methods. The best performance is highlighted in red (1st best) and blue (2nd best). RGDiffSR-i denotes the i-th epoch of the model.}
  \label{tradeoff}
\end{table}


Though the generated SR images exhibit commendable fidelity in Fig \ref{compare}, the model achieving the highest recognition accuracy seems comparatively weaker in traditional fidelity evaluation metrics like PSNR and SSIM. However, as depicted in Table \ref{tradeoff}, if RGDiffSR is trained to 1000 epochs, the model also manages to achieve a competitive PSNR/SSIM performance, albeit with a slight drop in accuracy. The improvement in PSNR/SSIM could be foreseen since the denoising network is optimized by pixel-wise loss functions. On the other hand, the decrease in accuracy indicates that higher fidelity in quantized metrics doesn't necessarily lead to higher recognition accuracy. This phenomenon is also observed in previous studies\cite{gestalt},\cite{C3}, indicating the inherent trade-off between fidelity and accuracy. 

\section{Conclusion}

In this paper, we propose a novel recognition-guided diffusion model for scene text image super-resolution (RGDiffSR). By integrating diffusion-based generative network and recognition guidance, RGDiffSR emerges as a cutting-edge solution for STISR, which exploits the inherent cross-modal generation capability of diffusion models, leading to improved fidelity and diversity in the generated images, especially in challenging scenarios. The recognition-guided denoising network serves as a critical component of RGDiffSR, expertly integrating semantic information into the diffusion process, which results in enhanced clarity and sharper character shapes in the generated images.
Experiments on the TextZoom dataset show that RGDiffSR consistently outperforms existing SOTA methods across various metrics, bringing generative methods back to the frontier of STISR again.
\bibliographystyle{IEEEbib}
\bibliography{strings,refs,icassp2024}

\end{document}